\documentclass[10pt,twocolumn,letterpaper]{article}

\usepackage[pagenumbers]{cvpr} 

\usepackage{graphicx}
\usepackage{amsmath}
\usepackage{amssymb}
\usepackage{booktabs}
\usepackage{caption,subcaption}
\usepackage[linesnumbered, ruled, vlined]{algorithm2e}

\SetCommentSty{mycommfont}

\makeatletter
\patchcmd{\@algocf@start}
{-1.5em}
{0pt}
{}{}
\makeatother

\graphicspath{{imgs/}}

\usepackage[pagebackref,breaklinks,colorlinks]{hyperref}

\usepackage[capitalize]{cleveref}
\crefname{section}{Sec.}{Secs.}
\Crefname{section}{Section}{Sections}
\Crefname{table}{Table}{Tables}
\crefname{table}{Tab.}{Tabs.}

\begin{document}
	\newcommand{\comm}[1]{}
	
	\title{Novelty Driven Evolutionary Neural Architecture Search}
	
	\author{Nilotpal Sinha\\
		National Yang Ming Chiao Tung University\\
		Hsinchu City, 30010, Taiwan\\
		{\tt\small nilotpalsinha.cs06g@nctu.edu.tw}
		\and
		Kuan-Wen~Chen\\
		National Yang Ming Chiao Tung University\\
		Hsinchu City, 30010, Taiwan\\
		{\tt\small kuanwen@cs.nctu.edu.tw}
	}
	\maketitle
	
	\begin{abstract}
		Evolutionary algorithms (EA) based neural architecture search (NAS) 
		involves evaluating each architecture by training it from scratch, which is
		extremely time-consuming. This can be reduced by using a supernet for
		estimating the fitness of an architecture due to weight sharing	among all
		architectures in the search space.
		However, the estimated fitness is very noisy due to the co-adaptation of the
		operations in the supernet which results in NAS methods getting trapped in
		local optimum. In this paper, we propose a method called NEvoNAS wherein the
		NAS problem is posed as a multi-objective problem with 2 objectives:
		\textit{(i)} maximize architecture novelty, \textit{(ii)} maximize architecture
		fitness/accuracy. The novelty search is used for maintaining a diverse set
		of solutions at each generation which helps avoiding local optimum traps while
		the architecture fitness is calculated using supernet. NSGA-II is used
		for finding the \textit{pareto optimal front} for the NAS problem and the best
		architecture in the pareto front is returned as the searched architecture. 
		Exerimentally, NEvoNAS gives better results on 2 different search spaces while
		using significantly less computational resources as compared to previous
		EA-based methods. The code for our paper can be found
		\href{https://github.com/nightstorm0909/NEvoNAS}{here}.
	\end{abstract}
	
	\section{Introduction}
	\label{sec:intro}
	In the recent years, convolutional neural networks (CNNs) have been very
	instrumental in solving various computer vision problems. However, the CNN
	architectures (such as ResNet \cite{he2016deep}, DenseNet \cite{huang2017densely}
	AlexNet \cite{krizhevsky2012imagenet}, VGGNet \cite{simonyan2014very})
	have been designed mainly by humans, relying on their intuition and
	understanding of the specific problem. Searching the neural architecture
	automatically by using an algorithm, i.e. \textit{Neural architecture search}
	(NAS), is an alternative to the architectures designed by humans, and in the recent
	years, these NAS methods have attracted increasing interest because of its promise
	of an automatic and efficient search of architectures specific to a task.
	Vanilla NAS methods
	\cite{elsken2018neural}\cite{zoph2016neural}\cite{zoph2018learning} have shown
	promising results in the field of computer vision but most of these methods consume
	a huge amount of computational power as it involves training each architecture
	from scratch for its evaluation. Vanilla evolutionary algorithm (EA)-based NAS
	methods also suffers from the same huge computational requirement problem. For
	example, the method proposed in \cite{real2019regularized} required 3150 GPU days
	of evolution.
	
	Recently proposed gradient-based methods such as \cite{liu2018darts2}
	\cite{dong2019searching}\cite{xie2018snas}\cite{dong2019one}
	\cite{chen2019progressive} have reduced	the search time	by sharing weights among
	the architectures through the use of supernet. A supernet represents all possible
	architectures in the search space while sharing the weights among all the
	architectures. However, the supernet suffers from inaccurate performance estimation
	which was first reported in \cite{bender2018understanding} and they showed that the
	co-adaptation among the operations in the compound edge leads to low correlation
	between the predicted performance via supernet and the true architecture
	performance from training-from-scratch. This results in premature convergence to
	the local optimum as shown in \cite{chen2019progressive} and
	\cite{Zela2020Understanding}. In order to mitigate this problem, NAS methods using
	supernet run the algorithm multiple times and select the best architecture out of
	the multiple runs. This can be thought of as running the algorithm multiple times 
	in order to get a set of good quality neural architectures. This is illustrated in
	Table~\ref{table:NAS_trials}, which shows the quality of the searched architecture
	in terms of test accuracy on CIFAR-10 dataset for gradient based method 
	DARTS\cite{liu2018darts2}, EA-based method EvNAS\cite{sinha2021evolving} and random
	search\cite{li2019random} in 4 trials. These multiple trials end up increasing the
	compuational costs. 
	
	\begin{table}[t]
		\centering
		\begin{tabular}{|c|c|c|c|c|}
			\hline
			NAS Methods& Trial 1 & Trial 2 & Trial 3 & Trial 4\\
			\hline
			DARTS$^{\dagger}$\cite{liu2018darts2} & $97.08$&$97.23$&$97.0$ & $96.95$\\	
			\hline
			EvNAS$^{\ddag}$\cite{sinha2021evolving} & $97.19$&$97.39$&$96.93$& $97.04$\\		
			\hline
			Random Search$^{\dagger}$\cite{li2019random} & $97.04$&$96.67$&$97.17$& $97.0$\\				
			\hline
		\end{tabular}
		\caption{Quality of the architectures found in 4 trials for different NAS
			methods using supernet. $^{\dagger}$ represents results report in
			\cite{li2019random} while $^{\ddag}$ represents re-run.
		}
		\label{table:NAS_trials}
	\end{table}	
	
	In this paper, we propose a method called NEvoNAS (\textit{Novelty Driven
		Evolutionary Neural Architecture Search}), in which the algorithm is run only
	once to get a set of good quality neural architecture solutions. This is achieved by
	posing the NAS problem as a multi-objective problem with 2 objectives: \textit{(i)}
	maximize architecture novelty, and \textit{(ii)} maximize architecture
	fitness/accuracy. Maximizing architecture novelty (i.e. \textit{novelty search})
	is used for maintaining a diverse set of solutions at each generation which helps
	avoiding local optimum traps while maximizing the architecture fitness using
	supernet guides the search towards potential solutions. We used NSGA-II for
	finding	the	\textit{pareto optimal front} of the multi-objective NAS problem and
	the best architecture in the discovered pareto optimal front is returned as the
	searched architecture. 
	
	Our contributions can be summarized as follows:
	\begin{itemize}
		\item We propose a novelty metric called \textit{architecture novelty metric}
		which determines how novel a neural architecture is from the already discovered
		neural architectures.
		
		\item We pose the NAS problem as a multi-objective problem with the objective
		of maximizing both the architecture novelty metric and architecture fitness.
		
		\item We also created a visualization of the search performed by NEvoNAS to get
		insights into the search process.
	\end{itemize}
	
	\section{Background}
	\subsection{Neural Architecture Search}
	\comm{Early NAS approaches\cite{stanley2002evolving}\cite{stanley2009hypercube},
		optimized both the neural architectures and the weights of the network using
		evolution. However,	their usage was limited to shallow networks. Recent
		\cite{zoph2016neural}\cite{pmlr-v80-pham18a}\cite{real2019regularized}
		\cite{zoph2018learning}\cite{real2017large}\cite{liu2018hierarchical}
		\cite{xie2017genetic} perform the architecture search separately while using
		gradient descent for optimizing the weights of the architecture for its
		evaluation which has made the search of deep networks possible.}
	Neural architecture search (NAS) methods can be classified into two categories:
	\textit{gradient-based} methods	and \textit{non-gradient based} methods.
	
	\subsubsection{Gradient-Based Methods:} \comm{These methods begin with a random neural
		architecture and the neural architecture is then updated using the gradient
		information on the basis of its performance on the validation data.}
	In general,	these methods, \cite{liu2018darts2}\cite{dong2019searching}
	\cite{xie2018snas}\cite{dong2019one}, relax the discrete architecture search
	space to a continuous search space by using a supernet. The	performance of the
	supernet on the validation data is used for updating the architecture using
	gradients. As the supernet shares weights among all architectures
	in the search space, these methods take	lesser time in the evaluation process
	and thus shorter search time. However, these methods suffer from the overfitting
	problem	wherein the	resultant architecture shows good performance on the validation
	data but exhibits poor performance on the test data. This can be attributed to its
	preference for parameter-less operations in the search space, as it leads to rapid
	gradient descent, \cite{chen2019progressive}.
	In contrast to these gradient-based methods, our method does not suffer from the
	overfitting problem because of its stochastic nature.
	
	\subsubsection{Non-Gradient Based Methods:}
	These methods include reinforcement learning (RL) methods and evolutionary
	algorithm (EA) methods. In the RL methods \cite{zoph2016neural}
	\cite{zoph2018learning}, an agent is used for the generating neural architecture
	and the agent is then trained to generate architectures in order to maximize
	its	expected accuracy on the validation	data. These accuracies were calculated by
	training the architectures from	scratch to convergence which resulted in long
	search time. This was improved in \cite{pmlr-v80-pham18a} by using a single
	directed acyclic graph (DAG) for sharing the weights among all the sampled
	architectures, thus resulting in reduced computational resources.
	The EA based NAS methods begin with a population of architectures and each
	architecture in the population is evaluated on the basis of its performance on the
	validation data. The popluation is then evolved	on the basis of the performance of
	the population. Methods such as those proposed in \cite{real2019regularized} and
	\cite{xie2017genetic} used gradient descent for	optimizing the weights of each
	architecture in the population from scratch in order to determine their accuracies 
	on the validation data as their fitness, resulting in huge computational
	requirements. In order to speed up the training process, in \cite{real2017large},
	the authors introduced weight inheritance wherein the architectures in the
	new generation population inherit the weights of the previous generation
	population, resulting in bypassing the training from scratch. However, the speed
	up gained is less as it still needs	to optimize the weights of the architecture.
	Methods such as that proposed in \cite{sun2019surrogate} used a random forest for
	predicting the performance of the architecture during the evaluation process,
	resulting in a high speed up as	compared to previous EA methods. However, its
	performance was far from the state-of-the-art results. In contrast, our method
	achieved better results than previous EA methods while using significantly less
	computational resources.
	
	\subsection{Novelty Search}
	Novelty search\cite{lehman2008exploiting} is an exploratory algorithm which is
	driven by the novelty of a behavior in the search space. This results in the
	search process exploring notably different areas of the search space which
	effectively	helps in avoiding the local optimum. In \cite{lehman2011evolving},
	the authors used novelty search for maintaining a diverse set of solutions in
	evolving virtual creatures.	Early work on novelty search based EA
	\cite{lehman2011abandoning} has shown promising results in searching
	for smaller networks. In \cite{zhang2020one}, novelty search was used as a
	controller to sample architectures for supernet training and the final result
	reported for the CIFAR-10 dataset was out of 10 independent runs/trials. In contrast,
	our	method runs once in order to get a set of possible solutions and uses novelty
	search for both supernet training and maintaining diverse set of solutions during
	the architecture search.
	
	\begin{figure}[h]
		\centering
		\begin{center}
			\includegraphics[width=0.8\linewidth]{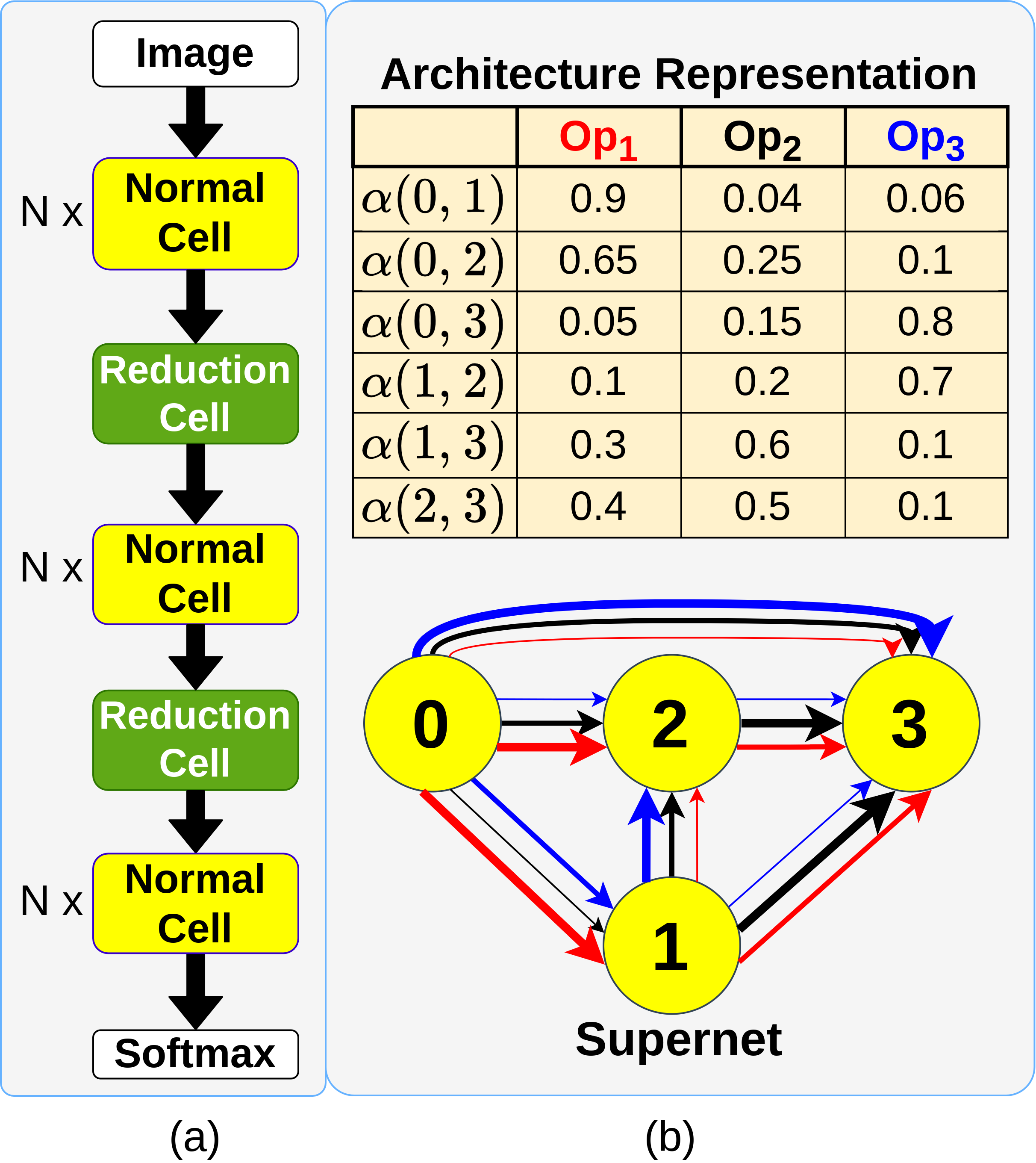}
		\end{center}
		\caption{
			(a) Architecture created by staking cells.
			(b) Architecture representation for a search space with 3 operations
			and 4 nodes. The thickness of the arrow in the supernet is proportional
			to the weight given to an operation.
		}
		\label{fig:architecture_representation}
	\end{figure}
	
	\section{Proposed Method}
	\subsection{Search Space and Architecture Representation}
	\label{subsect:search_space}	
	We follow \cite{pmlr-v80-pham18a}\cite{real2019regularized}\cite{zoph2018learning}
	\cite{liu2018darts2}\cite{dong2019searching}\cite{dong2019one}\cite{lu2020multi}
	\cite{liu2018progressive} to create the architecture by staking together 2 types 
	of cells: \textit{normal} cells which preserve the dimentionality of the
	input with a stride of one and \textit{reduction} cells which reduce the spatial
	dimension with a stride	of two, shown in
	Figure~\ref{fig:architecture_representation}(a).
	As illustrated in Figure~\ref{fig:architecture_representation}(b), a cell in the
	architecture is	represented	by an \textit{architecture parameter}, $\alpha$.
	Each $\alpha$ for a	normal cell and a reduction cell is represented by a matrix
	with columns representing the weights of different operations $Op(.)s$ from the
	operation space	$\mathcal O$ (i.e. the search space of NAS) and rows representing
	the edge between two nodes. For example, in
	Figure~\ref{fig:architecture_representation}(b), $\alpha(1, 2)$ represents the edge
	between node 1 and node 2 and the entries in the row represent the weights given
	to the three different operations.
	
	\comm{
		\begin{figure}[h]
			\centering
			\begin{center}
				\includegraphics[width=\linewidth]{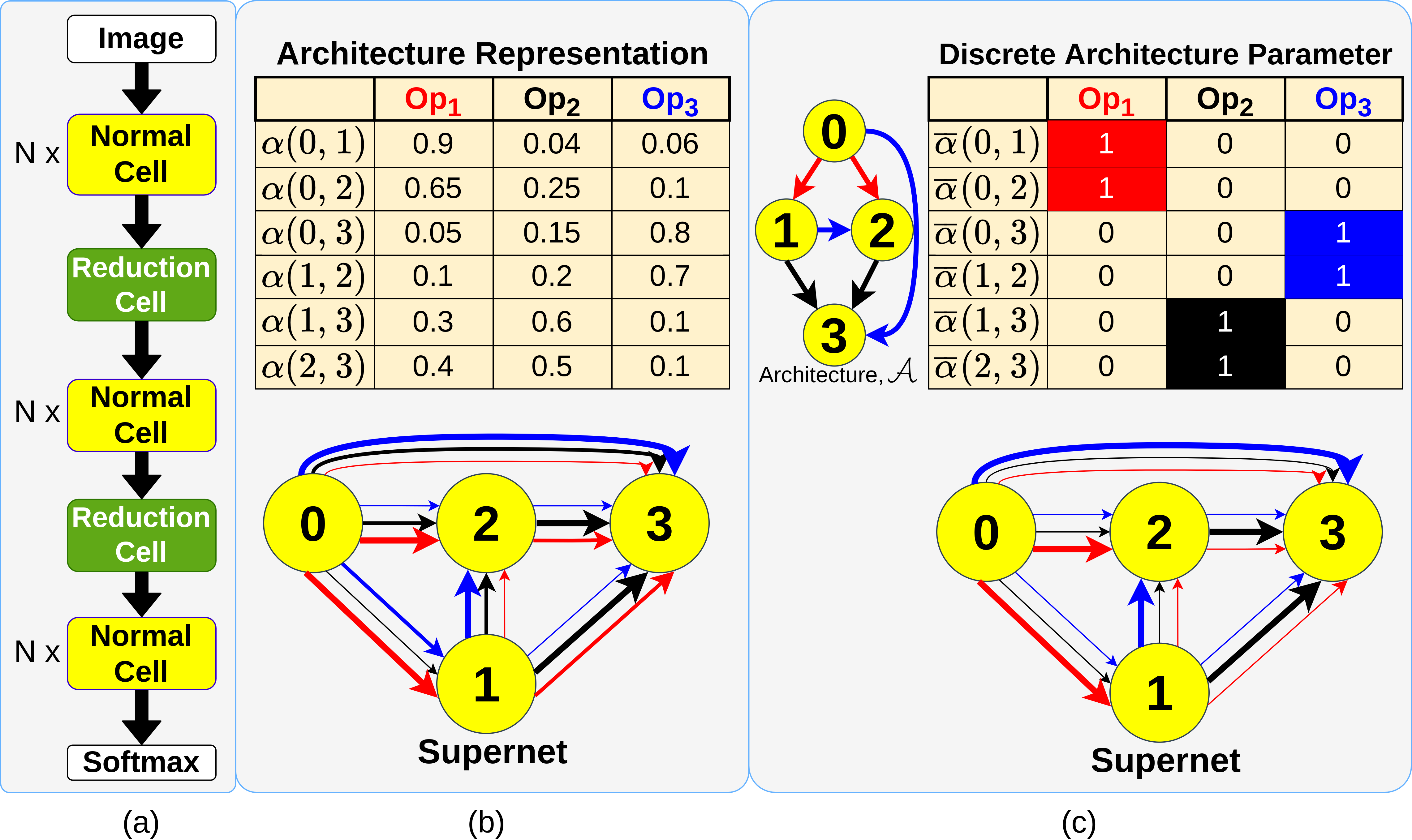}
			\end{center}
			\caption{
				(a) Architecture created by staking cells.
				(b) Architecture representation for a search space with 3 operations
				and 4 nodes. The thickness of the arrow in the supernet is proportional
				to the weight given to an operation.
				(c) Illustration of selecting an architecture in the supernet. The
				highlighted cell represents the selected operation between any two nodes.
			}
			\label{fig:arch_representation}
		\end{figure}
	}
	
	\begin{figure}[t]
		\centering
		\begin{center}
			\includegraphics[width=0.8\linewidth]{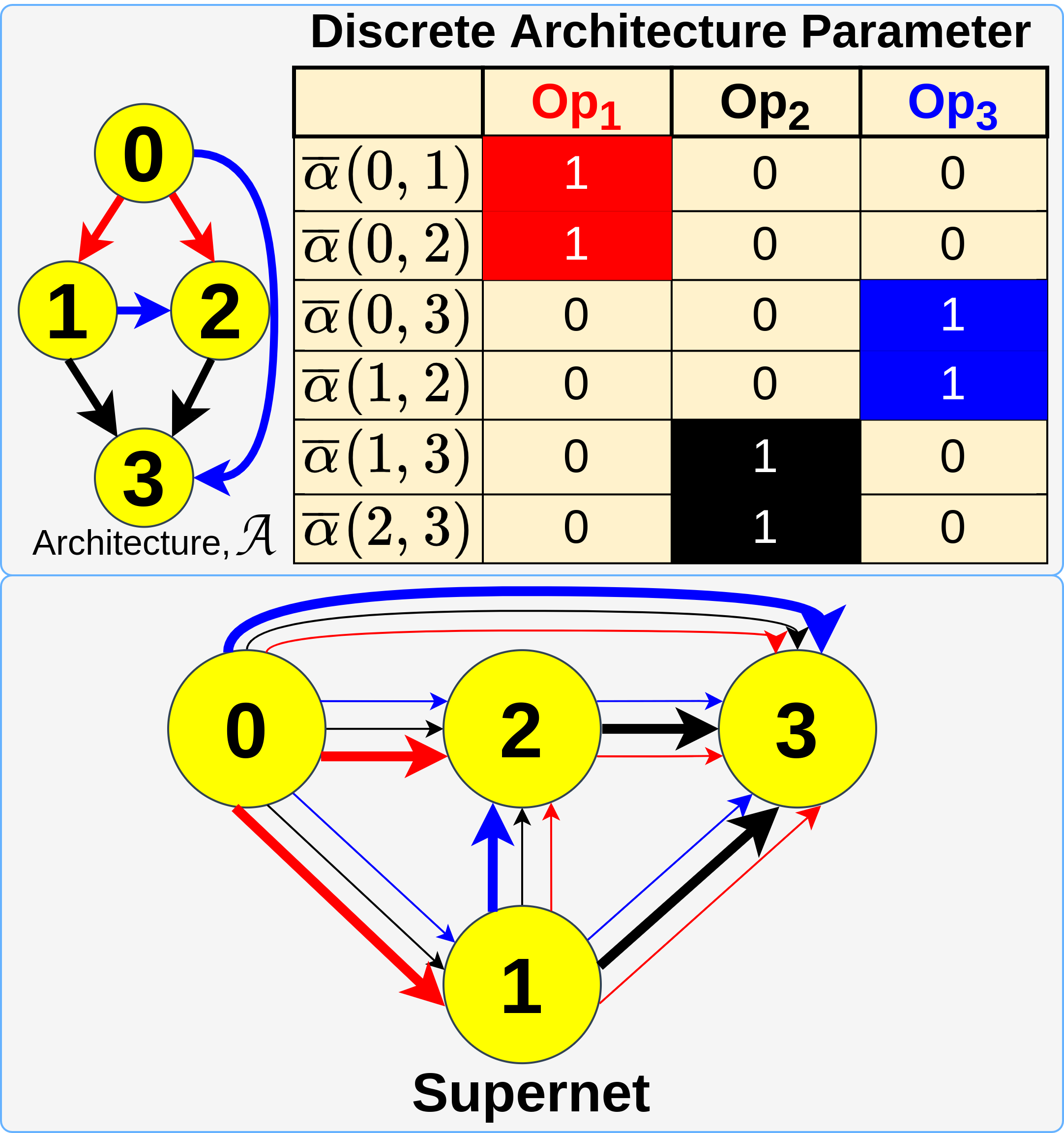}
		\end{center}
		\caption{
			Illustration of selecting an architecture, $\mathcal A$, in the supernet.
			The highlighted cell in the discrete architecture parameter, $\bar{\alpha}$,
			represents the selected operation between any two nodes. The thickness of the
			arrow in the supernet is proportional to the weight given to an operation.
		}
		\label{fig:arch_selection}
	\end{figure}
	
	\subsection{Performance Estimation}
	\label{subsect:performance}
	We used a \textit{supernet} \cite{liu2018darts2} to estimate the performance of
	an architecture in the search space. It shares the weights among all architectures
	in the search space	by treating all the architectures as the subgraphs
	of a supergraph. As illustrated in Figure~\ref{fig:architecture_representation}(b), 
	the	supernet uses the \textit{architecture parameter}, $\alpha$, wherein the
	directed edge from node $i$ to node $j$ is the weighted sum of all $Op(.)s$ in the
	operation space $\mathcal O$ (i.e. search space of NAS) where the $Op(.)s$ are
	weighted by the normalized $\alpha^{(i,j)}$ (normalized using softmax). This can
	be written as:
	\begin{equation}
		\label{eq:ons}
		f^{(i,j)}(x^{(i)}) = \sum_{Op \in \mathcal O } 
		\frac{exp(\alpha^{(i,j)}_{Op})}
		{\sum_{Op' \in \mathcal O } exp(\alpha^{(i,j)}_{Op'})} Op(x^{(i)})
	\end{equation}
	where $\alpha^{(i,j)}_{op}$ represents the weight of the operation $Op(.)$ in
	the operation space $\mathcal O$ between node $i$ and node $j$. This design
	choice allows us to skip the individual architecture training from
	scratch for its evaluation because of the weight-sharing nature of the supernet,
	thus resulting in a significant reduction of search time. 
	
	The performance	of an architecture is calculated using the supernet on
	the validation data, also known as the \textit{fitness} of the architecture. As
	illustrated in Figure~\ref{fig:arch_selection}, in order to select an
	architecture, $\mathcal A$, in the supernet, a new architecture parameter called
	\textit{discrete architecture} parameter, $\bar\alpha$, is created with the
	following entries:
	\begin{equation}
		\bar{\alpha}^{(i,j)}_{Op} = 
		\begin{cases}
			1, \text{if $Op(x^{(i)})$ present in $\mathcal A$}\\
			0,  \text{otherwise}\\
		\end{cases}    
	\end{equation}
	Using $\bar\alpha$, the architecture, $\mathcal A$, is selected in the supernet
	and the accuracy of the supernet on the validation data is used as the estimated
	\textit{fitness} of $\mathcal A$.
	\comm{This process gives higher equal
		weight to the architecture $\mathcal A$ operations while giving lower equal
		weights to the other architecture operations. This results in the
		higher contribution from the architecture $\mathcal A$ while very low
		contribution by other architectures during the fitness evaluation.
		This process can be thought of as selecting a subgraph (i.e. $\mathcal A$)
		from the supergraph	(i.e. supernet).
	}
	
	\begin{figure}[b]
		\centering
		\begin{center}
			\includegraphics[width=0.7\linewidth]{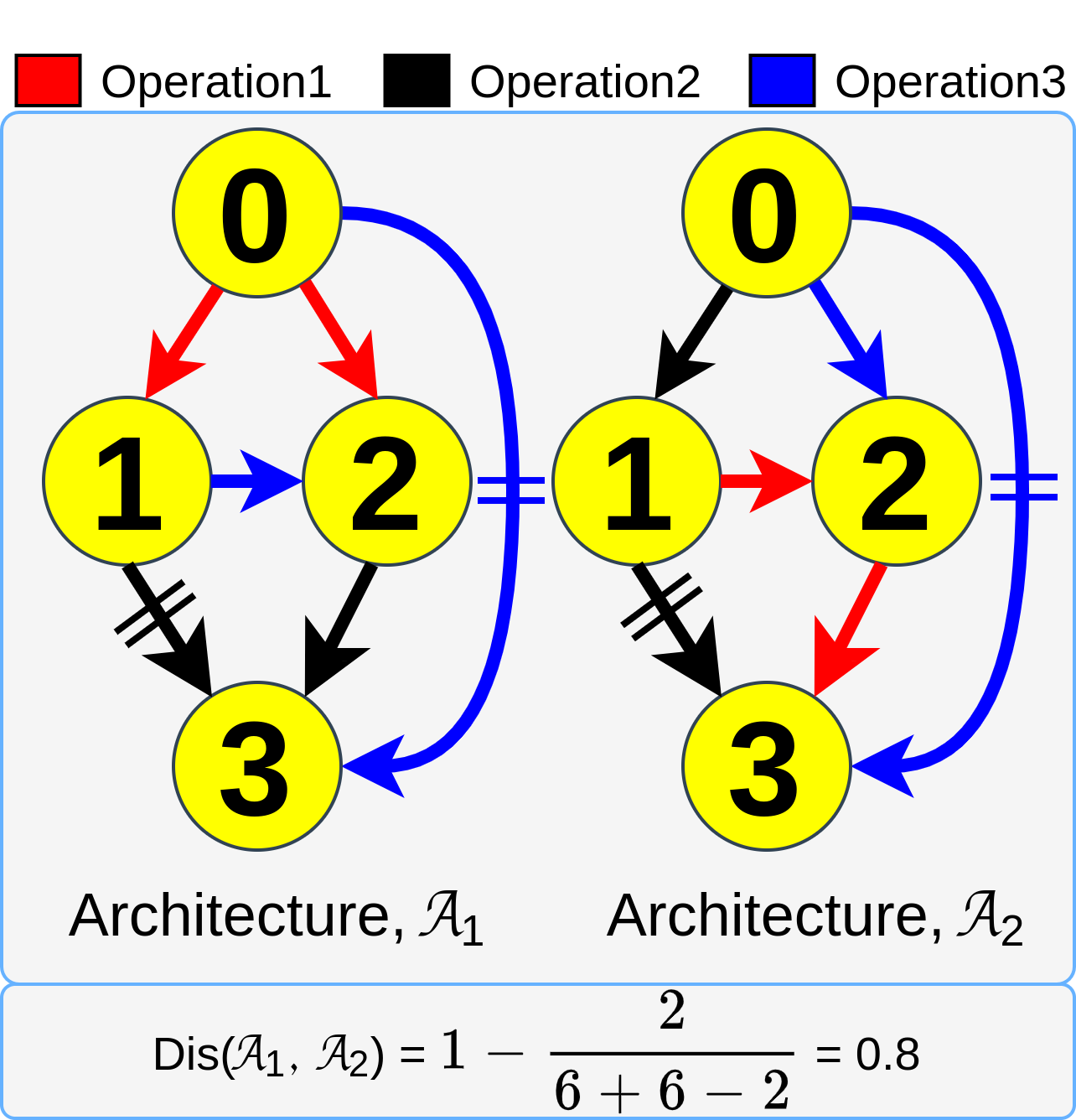}
		\end{center}
		\caption{
			Illustration of architecture dissimilarity metric calculation of two
			architectures $\mathcal A_1$ and $\mathcal A_2$. The common edges between
			$\mathcal A_1$ and $\mathcal A_2$ is shown by two small parallel lines on
			the	edges between the two nodes that are common in both architectures.
		}
		\label{fig:novelty_metric}
	\end{figure}
	
	\subsection{Architecture Novelty Metric}
	\label{subsec:novelty_metric}
	In order to create an EA algorithm that rewards novel architecture, we need a
	\textit{novelty metric} that measures how different an architecture is from
	another architecture. This provides a constant pressure to generate new
	architecture. In the neural architecture space, we first define a 
	\textit{similarity metric}, $Sim(\mathcal A_1, \mathcal A_2)$, which measures how
	similar an architecture $\mathcal A_1$ to another architecture $\mathcal A_2$ and
	is given as follows:
	\begin{equation}
		\label{eq:arch_similarity}
		Sim(\mathcal A_1, \mathcal A_2) = \frac{\cap({\mathcal A_1, \mathcal A_2})}
		{n(\mathcal A_1) + n(\mathcal A_2) - \cap({\mathcal A_1, \mathcal A_2})}
	\end{equation}
	Where $\cap({\mathcal A_1, \mathcal A_2})$ refers to the number of common
	operations between 2 nodes present in the given architectures,
	$\mathcal A_1, \mathcal A_2$, $n(\mathcal A_1)$ and $n(\mathcal A_2)$ refer
	to total number of opration edges present between nodes in the $\mathcal A_1$ and
	$\mathcal A_2$ respectively. Note that $Sim(\mathcal A_1, \mathcal A_2)$ equals to
	1 if $\mathcal A_1$ and $\mathcal A_2$ are the same architecture (i.e.
	$\mathcal A_1=\mathcal A_2$) and $Sim(\mathcal A_1, \mathcal A_2)$ equals to 0 if
	$\mathcal A_1$ and $\mathcal A_2$ do not share any operations between 2 nodes
	(i.e. completely different architectures). Thus,
	$0\le Sim(\mathcal A_1, \mathcal A_2)\le 1$. Now, we define an
	\textit{dissimilarity metric}, $Dis(\mathcal A_1, \mathcal A_2)$, which is used
	for measuring how different an architecture $\mathcal A_1$ is from another
	architecture $\mathcal A_2$ and is given as follows:
	\begin{equation}
		\label{eq:arch_dissimilarity}
		Dis(\mathcal A_1, \mathcal A_2) =  1 - Sim(\mathcal A_1, \mathcal A_2)
	\end{equation}
	Note that $Dis(\mathcal A_1, \mathcal A_2)$ equals to 0 if $\mathcal A_1$ and
	$\mathcal A_2$ are the same architecture (i.e. $\mathcal A_1=\mathcal A_2$) and
	$Dis(\mathcal A_1, \mathcal A_2)$ equals to 1 if $\mathcal A_1$ and
	$\mathcal A_2$ do not share any operations between 2 nodes (i.e. completely
	different architectures). Thus, $0\le Dis(\mathcal A_1, \mathcal A_2)\le 1$. For
	illustration, in Figure~\ref{fig:novelty_metric}, the architectures $\mathcal A_1$
	and $\mathcal A_2$ have two common edges between nodes $(0, 3)$ and $(1, 3)$, 
	thus $\cap({\mathcal A_1, \mathcal A_2}) = 2$,	while both $n(\mathcal A_1)$
	and $n(\mathcal A_2)$ are equal to 6. So, the dissimilarity metric comes out
	to be 0.8.
	
	The novelty of a newly generated neural architecture is computed with respect to
	an archive of past generated neural architectures and current population of
	neural architecture. To get the novelty of neural architecture, we need a novelty
	metric \cite{lehman2011evolving} which characterizes how far the neural
	architecture is from its predecessors and the rest of the population in the neural
	architectural space. We define \textit{architecture novelty metric} as the mean
	dissimilarity metric of the k-nearest neighbors, which is given as follows:
	\begin{equation}
		\label{eq:arch_novelty}
		F_{nov}(\mathcal A) =  \frac{1}{k}
		\sum_{i=1}^{k} Dis(\mathcal A, \mathcal A_i)
	\end{equation}
	Where $\mathcal A_i$ is the $i$th-nearest neighbor of the neural architecture
	$\mathcal A$ in terms of the dissimilarity metric. The nearest neighbors are
	calculated from the archive of past neural architectures and the current
	population.
	
	\begin{algorithm}[h]
		\caption{NEvoNAS}	
		\label{algo:NEvoNAS}
		\SetAlgoLined
		\KwIn{Population size $N_{pop}$, 
			training data $\mathcal{D}_{tr}$, total number of training batches
			$\mathcal B$,
			validation data $\mathcal{D}_{va}$,
			number of generations $\mathcal{G}$.
		}
		\KwOut{Pareto optimal front, $P_{optimal}$.}
		$P \gets$ Initialize population for NSGA-II algorithm\;
		$S \gets$ Initialize the supernet\;
		$archive \gets$ Initialize to empty set\;
		\For{ $g = 1, 2,..., \mathcal{G}$}
		{
			\tcc{Training supernet using current population}
			\For{ $i = 1, 2,..., \mathcal{B}$}
			{
				Copy $\alpha$[$i$ mod $N_{pop}$] to $S$\;
				Train $S$ on training batch[$i$]\;
			}
			\tcc{Evaluate architecture using the supernet}
			$F_{acc} \gets$ EvaluatePopulation($S$, $P$, $\mathcal{D}_{va}$)\;
			\tcc{Evaluate novelty of architectures in population}
			$F_{nov} \gets$ CalculateNovelty($P$, $archive$)\;
			UpdateArchive($P$, $archive$)\;
			\tcc{Apply NSGA-II to get next generation population}
			$P \gets$ NSGA-II($F_{acc}$, $F_{nov}$)\;
		}
	\end{algorithm}
	
	\subsection{NEvoNAS}
	\label{subsect:NEvoNAS}
	Multi-objective optimization is a popular branch of \textit{evolutionary
		computation} (EC), which involves optimizing problems with more than one
	objective function simultaneously.
	NEvoNAS poses the NAS problem as a multi-objective problem with two objectives:
	\textit{(i)} maximize architecture novelty, \textit{(ii)} maximize architecture
	fitness. The architecture novelty is calculated using the architecture novelty 
	metric (discussed in Section~\ref{subsec:novelty_metric}) and the fitness of the
	architecture is calculated using the supernet (discussed in
	Section~\ref{subsect:performance}). In order to solve the multi-objective problem,
	we used NSGA-II \cite{deb2002fast}, a well-known Pareto-based Multi-objective
	Evolutionary Algorithm (MOEA).
	
	The entire process is summarized in	Algorithm~\ref{algo:NEvoNAS}. NEvoNAS starts 
	with initializing the population randomly, the supernet with random weights and
	an empty $archive$. In each generation, the supernet is trained on the training
	data. During training, $\alpha$ of each individual architecture in the population
	is copied to the supernet in a round-robin fashion for each training batch. Then,
	the fitness of each individual architecture in the population,
	$F_{acc}$, is calculated using the supernet. Next, the novelty of each individual
	architecture in the population, $F_{nov}$, is calculated with respect to the
	$archive$ of past neural architectures and the current population of neural
	architectures. Th $archive$ is then updated to include the new individual
	architectures from the current population. NSGA-II is then used to generate the
	next generation	population. The entire process runs for $\mathcal G$ generations.
	NEvoNAS returns a pareto optimal front, $P_{optimal}$, (i.e. set of possible neural
	architecture solution) and the best neural architecture in the front is returned as
	the	searched architecture. Note that NEvoNAS runs for only once to get a set of
	possible solutions unlike other NAS methods using supernet
	\cite{liu2018darts2}\cite{sinha2021evolving}\cite{li2019random}.
	
	\section{Experiments}
	\label{experiments}
	\subsection{Search Spaces}
	In this section, we report the performance of NEvoNAS in terms of a neural 
	architecture search on two different search spaces: 1) \textit{Search
		space 1	(\textbf{S1})} \cite{liu2018darts2} and 2) \textit{Search space 2
		(\textbf{S2})}\cite{Dong2020NAS-Bench-201}. In \textit{S1}, we search for both
	normal and reduction cells where each node	$x^{(j)}$ maps two inputs to one
	output. Here, each cell has seven nodes with first two nodes being the output from
	previous cells and last node as output node, resulting in 14 edges among them.
	There are 8 operation in S1, so each architecture is represented by two 14x8
	matrices, one each for normal cell and reduction cell. So, an architecture,
	$\alpha$, is represented by a vector of size $2\times14\times8=224$ for S1. In
	\textit{S2}, we	search for only normal cells, where each node $x^{(j)}$ is
	connected to the previous node $x^{(i)}$ (i.e. $i < j$). It is a smaller search
	space where we only	search for the normal cell in
	Figure~\ref{fig:architecture_representation}(a). It provides a unified benchmark for
	almost any up-to-date NAS algorithm by providing results of each architecture
	in the search space on	CIFAR-10, CIFAR-100 and	ImageNet-16-12. Here, each cell
	has four nodes with first node as input node and last node as output node,
	resulting in 6 edges among them. There are 5 operations	in S2, so each
	architecture is represented by one 6x5 matrix for the normal cell. So, an 
	architecture, $\alpha$, is represented by a vector of size $6\times5=30$ for S2.
	
	\subsection{Dataset} \textbf{CIFAR-10} and \textbf{CIFAR-100}
	\cite{krizhevsky2009learning} has 50K training images and 10K testing images with
	images classified into 10 classes and 100 classes respectively. \textbf{ImageNet}
	\cite{imagenet_cvpr09} is well known benchmark for image classification
	containing 1K classes with 1.28 million training images and 50K images test images.
	\textbf{ImageNet-16-120} \cite{chrabaszcz2017downsampled} is a down-sampled variant
	of ImageNet where the original ImageNet is downsampled to 16x16 pixels with labels
	$\in\left[0,120\right]$ to construct ImageNet-16-120 dataset.
	The settings used for the datasets in \textbf{S1} are as follows:
	\begin{itemize}
		\item \textit{CIFAR-10}: We split 50K training images into two sets of size
		25K each, with one set acting as the training set and the other set as the
		validation set.
		\item \textit{CIFAR-100}: We split 50K training images into two sets. One set
		of size 40K images becomes the training set and the other set of size 10K
		images becomes the validation set.
	\end{itemize}
	The settings used for the datasets in \textbf{S2} are as follows:
	\begin{itemize}
		\item \textit{CIFAR-10}: The same settings as those used for S1 is used here
		as well.
		\item \textit{CIFAR-100}: The 50K training images remains as the training set
		and the 10K testing images are split into two sets of size 5K each, with one
		set acting as the validation set and the other set as the test set.
		\item \textit{ImageNet-16-120}: It has 151.7K training images, 3K validation
		images and 3K test images.
	\end{itemize}
	
	\begin{table}[b]
		\caption{Comparison of NEvoNAS with other NAS methods in S1 in terms
			of test accuracy (higher is better) on CIFAR-10.}
		\label{table:c10_s1}
		\begin{subtable}{\textwidth}
			\scalebox{0.85}{
				\begin{tabular}{l|c|c|c|c}
					
					\hline
					\multicolumn{1}{c|}{\bf{Architecture}} & \multicolumn{1}{c|}{\bf{Top-1}} & \bf{Params} & \bf{GPU} &\bf{Search} \\
					&\textbf{Acc.} (\%)&(M)& \bf{Days} & \bf{Method}\\
					\hline
					ResNet\cite{he2016deep}  & 95.39 & 1.7 & - & manual\\
					DenseNet-BC\cite{huang2017densely}& 96.54 & 25.6 & - & manual\\
					ShuffleNet\cite{zhang2018shufflenet}& 90.87 & 1.06 & - & manual\\
					\hline
					
					PNAS\cite{liu2018progressive}  & 96.59 & 3.2 & 225 & SMBO\\
					RSPS\cite{li2019random}         & 97.14 & 4.3 &2.7 & random\\
					\hline
					
					NASNet-A\cite{zoph2018learning}     & 97.35 & 3.3  &1800 & RL\\
					ENAS\cite{pmlr-v80-pham18a}    & 97.14  & 4.6  &0.45 & RL\\
					\hline
					DARTS\cite{liu2018darts2}     & 97.24 & 3.3 &4 & gradient\\
					GDAS\cite{dong2019searching}& 97.07 & 3.4 &0.83 & gradient\\
					SNAS\cite{xie2018snas}       & 97.15 & 2.8&1.5 & gradient\\
					SETN\cite{dong2019one}       & 97.31 & 4.6 &1.8 & gradient\\
					\hline
					
					AmoebaNet-A\cite{real2019regularized} & 96.66 & 3.2 &3150& EA\\
					Large-scale Evo.\cite{real2017large} & 94.60 & 5.4 &2750 & EA\\
					Hierarchical Evo.\cite{liu2018hierarchical}  & 96.25 & 15.7 &300 & EA\\
					CNN-GA\cite{sun2020automatically}  & 96.78 & 2.9 &35 & EA\\
					CGP-CNN\cite{suganuma2017genetic}  & 94.02 & 1.7 &27 & EA\\
					AE-CNN\cite{sun2019completely}  & 95.7 & 2.0 &27 & EA\\
					NSGANetV1-A2\cite{lu2020multi}  & 97.35 & 0.9 &27& EA\\
					AE-CNN+E2EPP\cite{sun2019surrogate}  & 94.70 & 4.3 &7 & EA\\
					NSGA-NET\cite{lu2019nsga}  & 97.25 & 3.3 &4& EA\\
					E$N^2$AS\cite{zhang2020one}  & 97.39 & 3.1 &3& EA\\
					\hline
					\bf{NEvoNAS-C10A}   & \bf{97.46} & \bf{3.4} & \bf{0.35}&\bf{EA}\\
					NEvoNAS-C10B        & 97.37      & 2.7  & 0.34  & EA\\
					NEvoNAS-C10C        & 97.29      & 3.2  & 0.34  & EA\\
					\hline
				\end{tabular}
			}
		\end{subtable}
	\end{table}
	
	\begin{table}[h]
		\caption{Comparison of NEvoNAS with other NAS methods in S1 in terms
			of test accuracy (higher is better) on CIFAR-100.}
		\label{table:c100_s1}
		\begin{subtable}{\textwidth}
			\scalebox{0.85}{
				\begin{tabular}{l|c|c|c|c}
					\hline
					\multicolumn{1}{c|}{\bf{Architecture}} &  \multicolumn{1}{c|}{\bf{Top-1}} & \bf{Params} & \bf{GPU} &\bf{Search} \\
					&\textbf{Acc.} (\%)& (M) & \bf{Days} &\bf{Method}\\
					\hline
					ResNet\cite{he2016deep}  & 77.90 & 1.7 & - & manual\\
					DenseNet-BC\cite{huang2017densely} & 82.82   & 25.6 & - & manual\\
					ShuffleNet\cite{zhang2018shufflenet}& 77.14 & 1.06 & - & manual\\
					\hline
					PNAS\cite{liu2018progressive}  & 80.47 & 3.2 & 225 & SMBO\\
					\hline
					
					\comm{NASNet-A\cite{zoph2018learning}    & -     & 3.3  &1800& RL\\}
					MetaQNN\cite{baker2017designing}     & 72.86      & 11.2  &90& RL\\
					ENAS\cite{pmlr-v80-pham18a}     & 80.57      & 4.6  &0.45& RL\\
					\hline
					
					
					AmoebaNet-A\cite{real2019regularized}  & 81.07 & 3.2 &3150& EA\\
					Large-scale Evo.\cite{real2017large}  & 77.00 & 40.4 &2750& EA\\
					CNN-GA\cite{sun2020automatically}  & 79.47 & 4.1 &40& EA\\
					AE-CNN\cite{sun2019completely}  & 79.15 & 5.4 &36& EA\\
					NSGANetV1-A2\cite{lu2020multi} & 82.58 & 0.9 &27& EA\\
					Genetic CNN\cite{xie2017genetic}  & 70.95 & - &17& EA\\
					AE-CNN+E2EPP\cite{sun2019surrogate}  & 77.98 & 20.9 &10& EA\\
					NSGA-NET\cite{lu2019nsga}  & 79.26 & 3.3 &8& EA\\
					
					\hline
					\bf{NEvoNAS-C100A}  & \bf{83.95} & \bf{3.9} & \bf{0.3} & \bf{EA}\\
					NEvoNAS-C100B       & 82.93   & 2.9      & 0.3 & EA\\
					NEvoNAS-C100C       & 82.83   & 2.7      & 0.3 & EA\\
					\hline
				\end{tabular}
			}
		\end{subtable}
	\end{table}
	
	\begin{table}[h]
		\caption{Comparison of NEvoNAS with other NAS methods in S1 in terms
			of test accuracy (higher is better) on ImageNet.}
		\label{table:imagenet_s1}
		\begin{subtable}{\textwidth}
			\scalebox{0.66}{
				\begin{tabular}{l|c|c|c|c|c|c}
					\hline
					\multicolumn{1}{c|}{\bf{Architecture}} & \multicolumn{2}{c|}{\bf{Test Accuracy (\%)}} & \bf{Params} &+$\times$& \bf{GPU} &\bf{Search} \\
					& \bf{top 1} & \bf{top 5} & (M) & (M) & \bf{Days} & \bf{Method} \\
					
					\hline
					MobileNet-V2, (\cite{sandler2018mobilenetv2})&72.0& 91.0 & 3.4 & 300 & - & manual\\ 
					\hline
					PNAS, (\cite{liu2018progressive})  &74.2& 91.9    & 5.1 & 588 & 225 & SMBO\\
					\hline
					
					NASNet-A, (\cite{zoph2018learning})    & 74.0 & 91.6      & 5.3 & 564 &1800& RL\\
					NASNet-B, (\cite{zoph2018learning})    & 72.8  & 91.3    & 5.3 & 488 &1800& RL\\
					NASNet-C, (\cite{zoph2018learning})    & 72.5  & 91.0     & 4.9 & 558 &1800& RL\\
					\hline
					
					DARTS, (\cite{liu2018darts2})& 73.3  & 91.3 & 4.7 & 574 &4& gradient\\
					GDAS, (\cite{dong2019searching})  & 74.0  & 91.5 & 5.3 & 581 &0.83& gradient\\
					SNAS, (\cite{xie2018snas})& 72.7  & 90.8 & 4.3 & 522 &1.5& gradient\\
					SETN, (\cite{dong2019one})& 74.3  & 92.0 & 5.4 & 599 &1.8& gradient\\
					\hline
					
					AmoebaNet-A, (\cite{real2019regularized}) & 74.5 & 92.0 & 5.1 & 555 &3150& EA\\
					AmoebaNet-B, (\cite{real2019regularized}) & 74.0 & 91.5 & 5.3 & 555 &3150& EA\\
					AmoebaNet-C, (\cite{real2019regularized}) & 75.7 & 92.4 & 6.4 & 570 &3150& EA\\
					NSGANetV1-A2, (\cite{lu2020multi}) & 74.5 & 92.0 & 4.1 & 466 &27& EA\\
					\comm{FairNAS-A          \cite{chu2019fairnas} & 75.3 & 92.4 & 4.6 & 388 &12& EA\\
						FairNAS-B          \cite{chu2019fairnas} & 75.1 & 92.3 & 4.5 & 345 &12& EA\\
						FairNAS-C          \cite{chu2019fairnas} & 74.7 & 92.1 & 4.4 & 321 &12& EA\\
						\hline
						
						EvNAS-A (Ours)                                     & 75.6 & 92.6 & 5.1 & 570 & 3.83 & EA\\
						EvNAS-B (Ours)                                     & 75.6 & 92.6 & 5.3 & 599 & 3.83 & EA\\
						EvNAS-C (Ours)                                     & 74.9 & 92.2 & 4.9 & 547 & 3.83 & EA\\
					}
					\hline
					NEvoNAS-C10A & 74.8 & 92.1 & 4.9 & 541 & 0.35 & EA\\
					\bf{NEvoNAS-C100A}  & \bf{75.7} & \bf{92.7} & \bf{5.4} & \bf{598} & \bf{0.3} & \bf{EA}\\
					\hline
				\end{tabular}
			}
		\end{subtable}
	\end{table}
	
	\subsection{Implementation Details}
	\label{implementation_details}
	
	\begin{figure*}[t]
		\centering
		\subfloat[]{
			\includegraphics[width=0.45\linewidth]{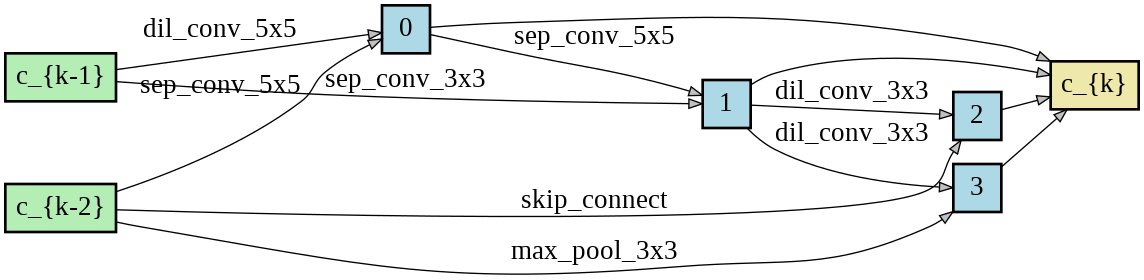}
		}
		\qquad
		\subfloat[]{
			\includegraphics[width=0.45\linewidth]{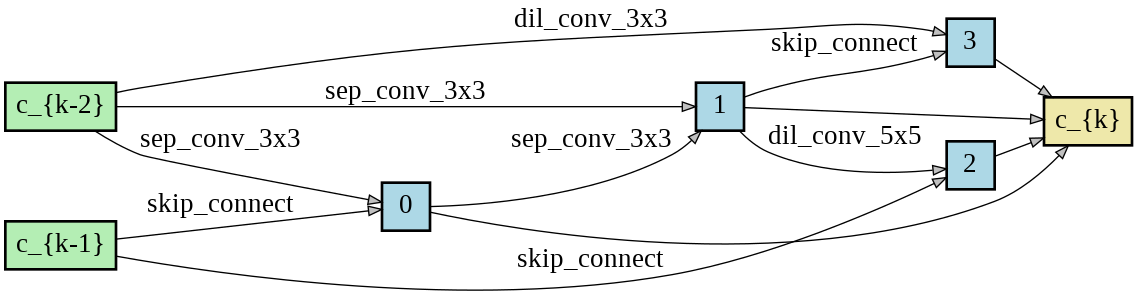}
		}
		\qquad
		\subfloat[]{
			\includegraphics[width=0.45\linewidth]{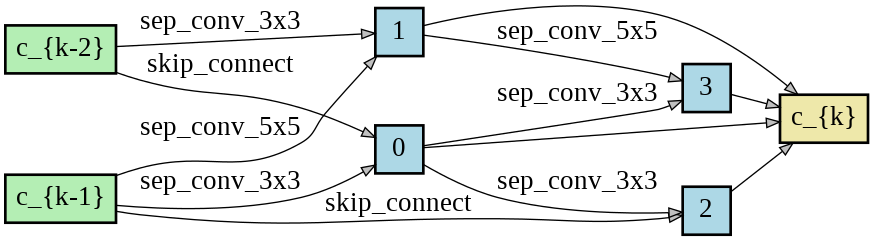}
		}
		\qquad
		\subfloat[]{
			\includegraphics[width=0.45\linewidth]{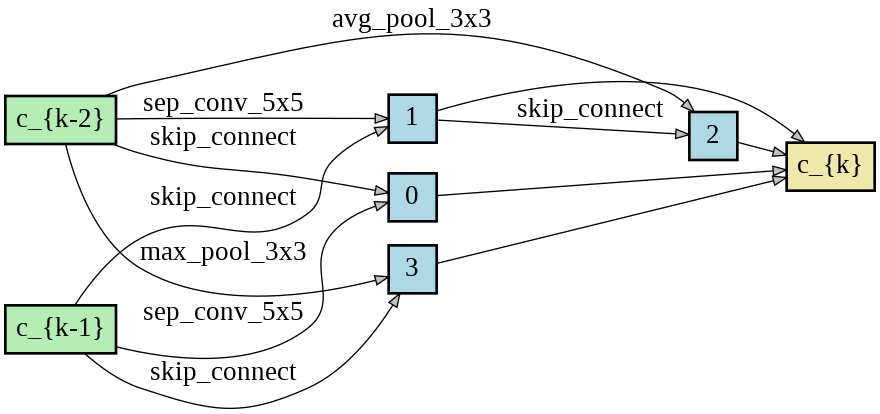}
		}		
		\caption{Cells discovered
			by NEvoNAS-C10A (a) Normal cell (b) Reduction cell;
			by NEvoNAS-C100A (c) Normal cell (d) Reduction cell.
		}
		\label{fig:searched_cells}
	\end{figure*}
	
	\subsubsection{\textbf{Evolutionary Algorithm Settings:}}
	In general, the supernet suffers from high memory requirements which makes it
	difficult to fit it in a single GPU. For S1, we follow \cite{liu2018darts2}
	and use a smaller supernet, called \textit{proxy model} which is created with 8
	stacked cells and 16 initial channels. All the other settings are also same for
	both datasets i.e. batch size of 64, weight decay $\lambda=3\times10^{-4}$,
	cutout\cite{devries2017improved}, initial learning rate $\eta_{max}=0.025$
	(annealed down to 0 by using a cosine schedule without
	restart\cite{DBLP:conf/iclr/LoshchilovH17}) and momentum $\rho=0.9$. For
	\textit{S2}, we do not use a proxy model as the size of the supernet is
	sufficiently small to be fitted in a single GPU. For training, we
	follow the same settings as those used in S1 for CIFAR-10, CIFAR-100 and
	ImageNet16-120 except batch size of 256.
	
	Following \cite{sun2019surrogate}\cite{sun2019completely}, the evolutionary
	algorithm (EA), for both \textit{S1} and \textit{S2}, uses a population size of 20
	in each	generation. For our NSGA-II	implementation, we used \textit{simulated 
		binary crossover} \cite{deb2007self} with $\eta=15$ and probability $=0.7$ for crossover,
	polynomial mutation \cite{deb2007self} with $\eta=20$ and probability $=0.1$ for mutation
	and the binary tournament for the selection method. The nearest-neighbors size for the
	novelty search is 5. Following \cite{liu2018darts2}\cite{sinha2021evolving}, NEvoNAS runs
	for 50 generations (i.e. $\mathcal G$) to get the pareto optimal front, $P_{optimal}$.
	We used	pymoo \cite{pymoo} (a python library) for the NSGA-II algorithm and pytorch
	\cite{NEURIPS2019_9015} for the supernet training and architecture evaluation on
	GPUs. All the above training and architecture search were performed on a single
	Nvidia RTX 3090 GPU.
	
	\subsubsection{\textbf{Architecture evaluation:}}
	The discovered architectures (i.e. discovered cells in the pareto optimal front,
	$P_{optimal}$) at the end of the architecture search are trained on the dataset to evaluate
	its performance. The best performing architecture within the pareto optimal front is
	returned as the	searched architecture of NEvoNAS and is used for comparing with
	other NAS methods. For \textit{S1}, we follow the training settings used in DARTS
	\cite{liu2018darts2}. Here, a larger network, called \textit{proxyless network} 
	\cite{li2019random}, is created using the discovered cells from the pareto optimal
	front with 20 stacked cells and 36 initial channels for both CIFAR-10 and
	CIFAR-100 datasets. It is then trained for 600 epochs on both the datasets with
	the same settings as the ones used in the supernet training above. Following
	recent works
	\cite{pmlr-v80-pham18a}\cite{real2019regularized}\cite{zoph2018learning}
	\cite{liu2018darts2}\cite{liu2018progressive}, we use an auxiliary tower with
	0.4 as its	weights, path dropout probability of 0.2 and cutout
	\cite{devries2017improved} for additional enhancements.
	For	ImageNet, the neural architecture is created with 14 cells and 48 initial
	channels in the	mobile setting, wherein the input image size is 224 x 224 and the
	number of multiply-add operations in the model is restricted to less than 600M. It
	is trained on 8 NVIDIA V100 GPUs by following the training settings used in
	\cite{chen2019progressive}.
	
	\begin{table*}[h]
		\caption{Comparison of NEvoNAS with other NAS methods on NAS-Bench-201 (i.e. 
			S2)\cite{Dong2020NAS-Bench-201} with mean $\pm$ std. accuracies on
			CIFAR-10, CIFAR-100 and ImageNet16-120 (higher is better).
			NEvoNAS (only Novelty) and NEvoNAS (only Accuracy) refer to applying
			NEvoNAS with single objective of novelty and architecture accuracy
			respectively. Optimal refers to the best architecture accuracy for each
			dataset. Search times are given for a CIFAR-10 search on a single GPU.}
		\label{table:NAS201}
		\centering
		\scalebox{0.8}{
			\begin{tabular}{l|c|cc|cc|cc|c}
				\hline
				\multicolumn{1}{c|}{\bf{Method}}& \bf{Search} & \multicolumn{2}{c|}{\bf{CIFAR-10}} & \multicolumn{2}{c|}{\bf{CIFAR-100}} & \multicolumn{2}{c|}{\bf{ImageNet-16-120}} & \bf{Search}\\
				&(seconds)& \it{validation} & \it{test} & \it{validation} & \it{test} & \it{validation} & \it{test}
				& \bf{Method}\\
				\hline
				RSPS \cite{li2019random} & $7587$ &$84.16\pm1.69$&$87.66\pm1.69$&$59.00\pm4.60$&$58.33\pm4.64$& $31.56\pm3.28$&$31.14\pm3.88$& random\\
				
				DARTS-V1 \cite{liu2018darts2} &$10890$ & $39.77\pm0.00$ & $54.30\pm0.00$ & $15.03\pm0.00$ &$15.61\pm0.00$ & $16.43\pm0.00$&$16.32\pm0.00$& gradient\\
				
				DARTS-V2 \cite{liu2018darts2} & $29902$ & $39.77\pm0.00$ & $54.30\pm0.00$ & $15.03\pm0.00$ &$15.61\pm0.00$& $16.43\pm0.00$&$16.32\pm0.00$& gradient\\
				
				GDAS \cite{dong2019searching} & $28926$ & $90.00\pm0.21$ & $93.51\pm0.13$ & $71.14\pm0.27$ &$70.61\pm0.26$& $41.70\pm1.26$&$41.84\pm0.90$& gradient\\
				
				SETN \cite{dong2019one} & $31010$ &$82.25\pm5.17$&$86.19\pm4.63$&$56.86\pm7.59$&$56.87\pm7.77$ &$32.54\pm3.63$&$31.90\pm4.07$& gradient\\
				
				ENAS \cite{pmlr-v80-pham18a} & $13314$ &$39.77\pm$0.00&$54.30\pm0.00$&$15.03\pm00$&$15.61\pm0.00$& $16.43\pm0.00$&$16.32\pm0.00$& RL\\
				
				EvNAS \cite{sinha2021evolving} & 22445 &88.98$\pm$1.40 & 92.18$\pm$1.11 & 66.35$\pm$2.59 & 66.74$\pm$3.08 & 39.61$\pm$0.72 & 39.00$\pm$0.44 & EA\\
				
				
				\bf{NEvoNAS} &\bf{9434}&\bf{90.58}$\pm$\bf{0.62}&\bf{93.50}$\pm$\bf{0.30}            &\bf{71.33}$\pm$\bf{0.88}&\bf{71.51}$\pm$\bf{0.52}& \bf{44.85}$\pm$\bf{0.46}&\bf{45.30}$\pm$\bf{0.82}&\bf{EA}\\
				
				NEvoNAS (only Novelty) & 28 &87.41$\pm$0.34 & 90.75$\pm$0.69 & 65.78$\pm$6.88 & 65.68$\pm$6.82 & 37.30$\pm$5.55 & 36.90$\pm$5.96 & EA\\
				
				NEvoNAS (only Accuracy) & 9706 &87.43$\pm$1.79 & 90.64$\pm$1.62 & 62.15$\pm$9.77 & 62.14$\pm$9.63 & 39.24$\pm$5.08 & 39.05$\pm$5.75 & EA\\
				
				\hline
				ResNet & N/A &$90.83$&$93.97$&$70.42$&$70.86$&$44.53$&$43.63$& manual\\
				Optimal & N/A &$91.61$&$94.37$&$73.49$&$73.51$&$46.77$&$47.31$& N/A\\
				\hline
				
		\end{tabular}}
	\end{table*}

	\begin{figure*}[h]
		\centering
		\begin{center}
			\includegraphics[width=\linewidth]{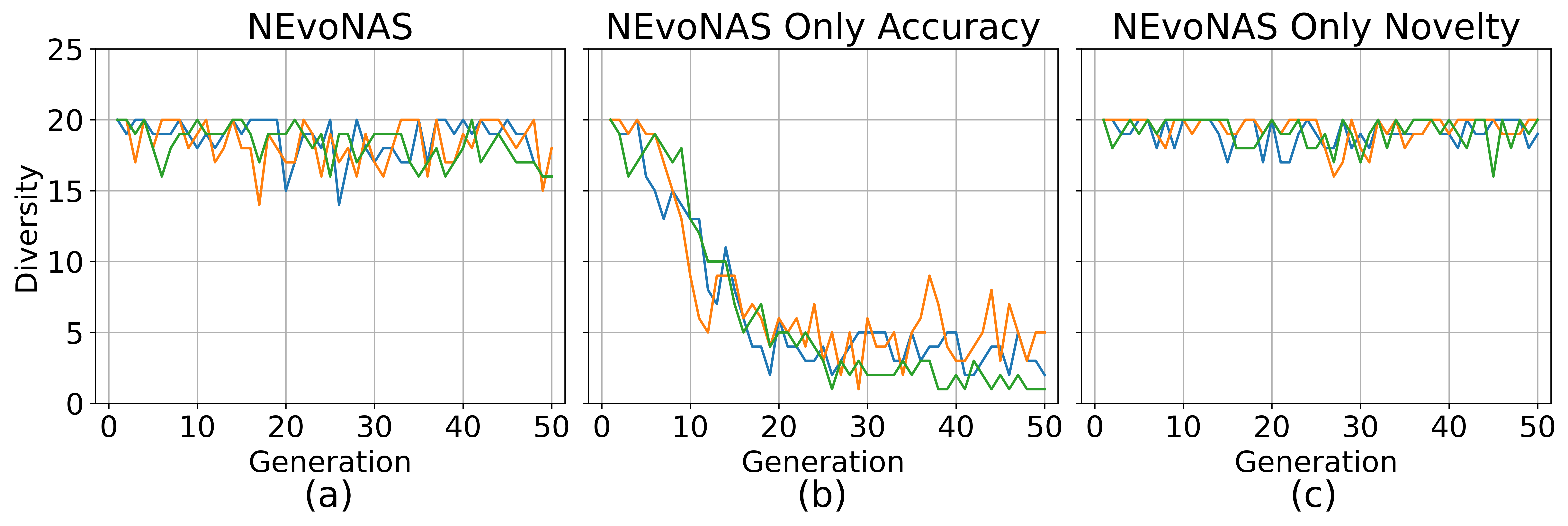}
		\end{center}
		\caption{Illustration of diversity in the population in each generation for
			the 3 independent runs of
			(a) NEvoNAS,
			(b) NEvoNAS using only accuracy as single objective,
			(c) NEvoNAS using only novelty as single objective.
		}
		\label{fig:diversity_comparision}
	\end{figure*}
	
	\begin{figure*}[h]
		\centering
		\begin{center}
			\includegraphics[width=\linewidth]{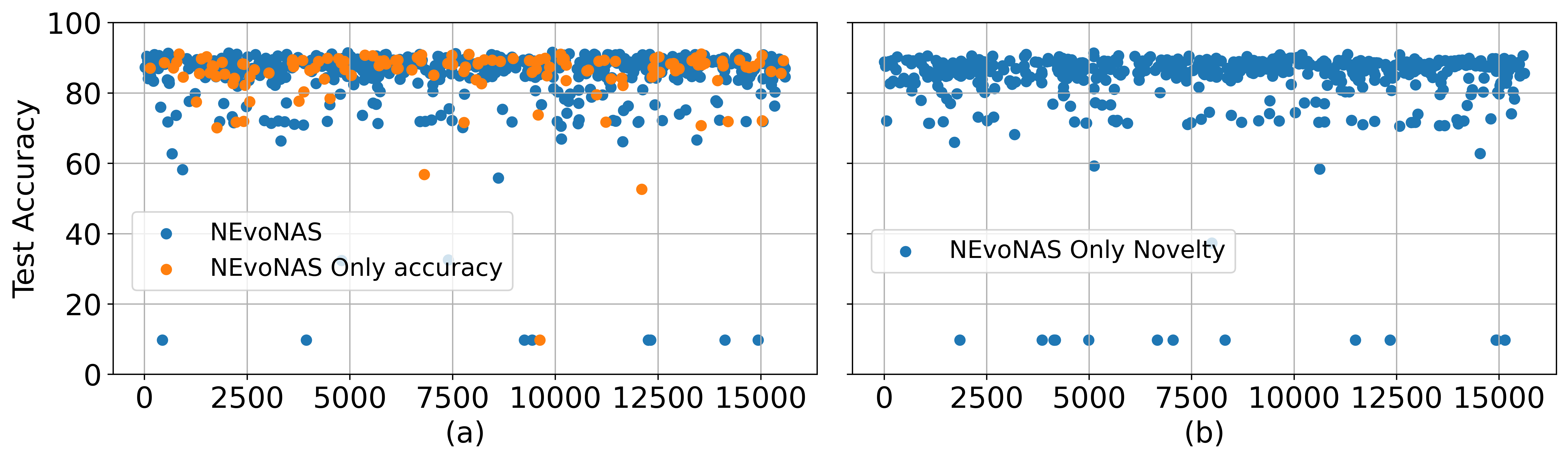}
		\end{center}
		\caption{Visualizing the architecture search space exploration by plotting the
			ground truth accuracies	of all the discovered architectures. The x-axis
			represents all 15,625 architectures in the search space S2
			\cite{Dong2020NAS-Bench-201}, the y-axis represents	the true test
			accuracies of the architectures and the dots represents	all the
			architectures discovered in a single run of
			(a) NEvoNAS and NEvoNAS using only accuracy,
			(b) NEvoNAS using only novelty.
		}
		\label{fig:search_exploration2}
	\end{figure*}
	
	\comm{
		\begin{figure}[t]
			\centering
			\begin{center}
				\includegraphics[width=\linewidth]{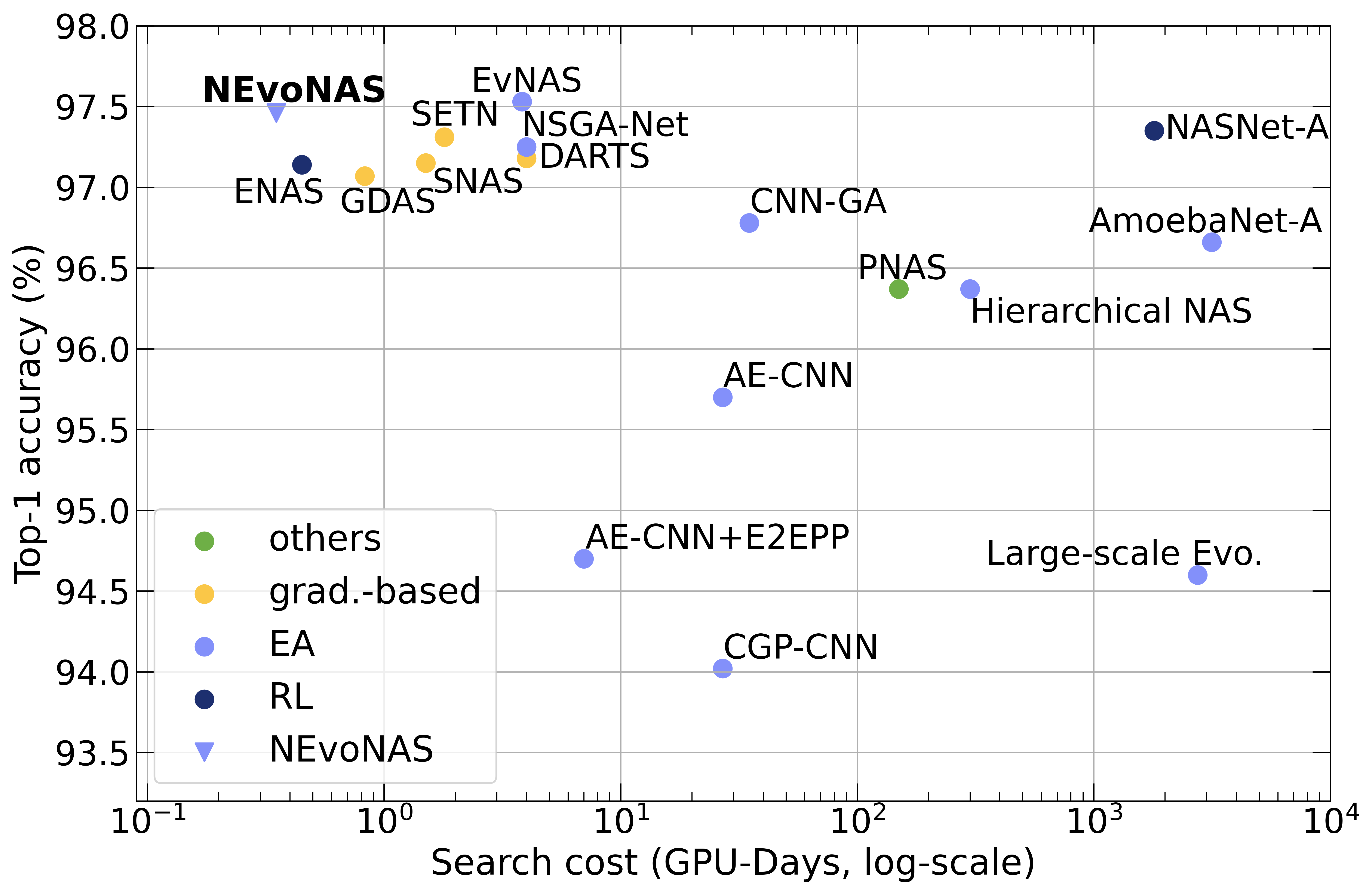}
			\end{center}
			\caption{Search cost
			}
			\label{fig:search_cost}
		\end{figure}
	}
	\subsection{Results}
	\label{results}
	\subsubsection{\textbf{Search Space 1 (S1):}}
	We performed 3 architecture searches on both CIFAR-10 and CIFAR-100 with
	different random number	seeds; their results are provided in
	Table~\ref{table:c10_s1} and Table~\ref{table:c100_s1}. The results show that
	the cells discovered by NEvoNAS on CIFAR-10 and CIFAR-100 achieve better results
	than those by human designed, RL based, gradient-based and EA-based methods.
	On comparing the computation time (or \textit{search cost}) measured in terms of
	\textit{GPU days}, we found that NEvoNAS performs the architecture search in
	significantly less time as compared to other EA-based methods while giving better
	search results. GPU days for any NAS method is calculated by multiplying the
	number of GPUs used in the NAS method by the execution time (reported in units
	of days). The best architectures discovered in single run on CIFAR-10 and
	CIFAR-100 (i.e. NEvoNAS-C10A and NEvoNAS-C100A) are shown in
	Figure~\ref{fig:searched_cells}. All the other discovered architectures for S1 are
	provided in the	supplementary.
	
	We followed	\cite{liu2018progressive}\cite{liu2018darts2}\cite{pmlr-v80-pham18a}
	\cite{real2019regularized}\cite{zoph2018learning} to
	compare the transfer capability of NEvoNAS with that of the other NAS methods,
	wherein the discovered architecture	on a dataset was transferred to another
	dataset (i.e. ImageNet) by retraining the architecture from scratch on the new
	dataset. So, the best discovered architectures from the architecture search on
	CIFAR-10 and CIFAR-100 (i.e. NEvoNAS-C10A and NEvoNAS-C100A) are then evaluated on
	the ImageNet dataset in mobile setting and the results are provided in
	Table~\ref{table:imagenet_s1}. The results show that the cells discovered by
	NEvoNAS on CIFAR-10 and CIFAR-100 can be successfully transferred to ImageNet,
	achieving better results than those of human designed, RL based, gradient based
	and EA based methods while using significantly less computational resources.
	
	\subsubsection{\textbf{Search Space 2 (S2):}}
	Following \cite{Dong2020NAS-Bench-201}, we performed 3 architecture searches
	with different random number seeds each on CIFAR-10, CIFAR-100 and ImageNet-16-120
	and compared the performance of	NEvoNAS with other NAS methods in 
	Table~\ref{table:NAS201}. The results show that NEvoNAS outperforms all of the NAS
	methods on all 3 datasets. We also performed the architecture search
	using NEvoNAS with only novelty as a single objective (i.e. novelty search,
	reported as NEvoNAS (only Novelty) in Table~\ref{table:NAS201}) and found that
	novelty search alone gives sub-optimal results with high variance as it is not
	searching for good quality architecture. Lastly, we performed the architecture
	search using NEvoNAS with only architecture accuracy/fitness using supernet as a
	single objective (reported as NEvoNAS (only Accuracy) in Table~\ref{table:NAS201})
	and found that it also gives sub-optimal results with high variance. Note that the
	high variance of NEvoNAS (only Accuracy) shows the need for running those NAS
	methods which use supernet, multiple times in order to select the best architecture
	out of the multiple runs. In contrast, our method provides a set of neural
	architecture solutions in a single run.
	
	\comm{
	\begin{figure}[h]
		\centering
		\begin{center}
			\includegraphics[width=\linewidth]{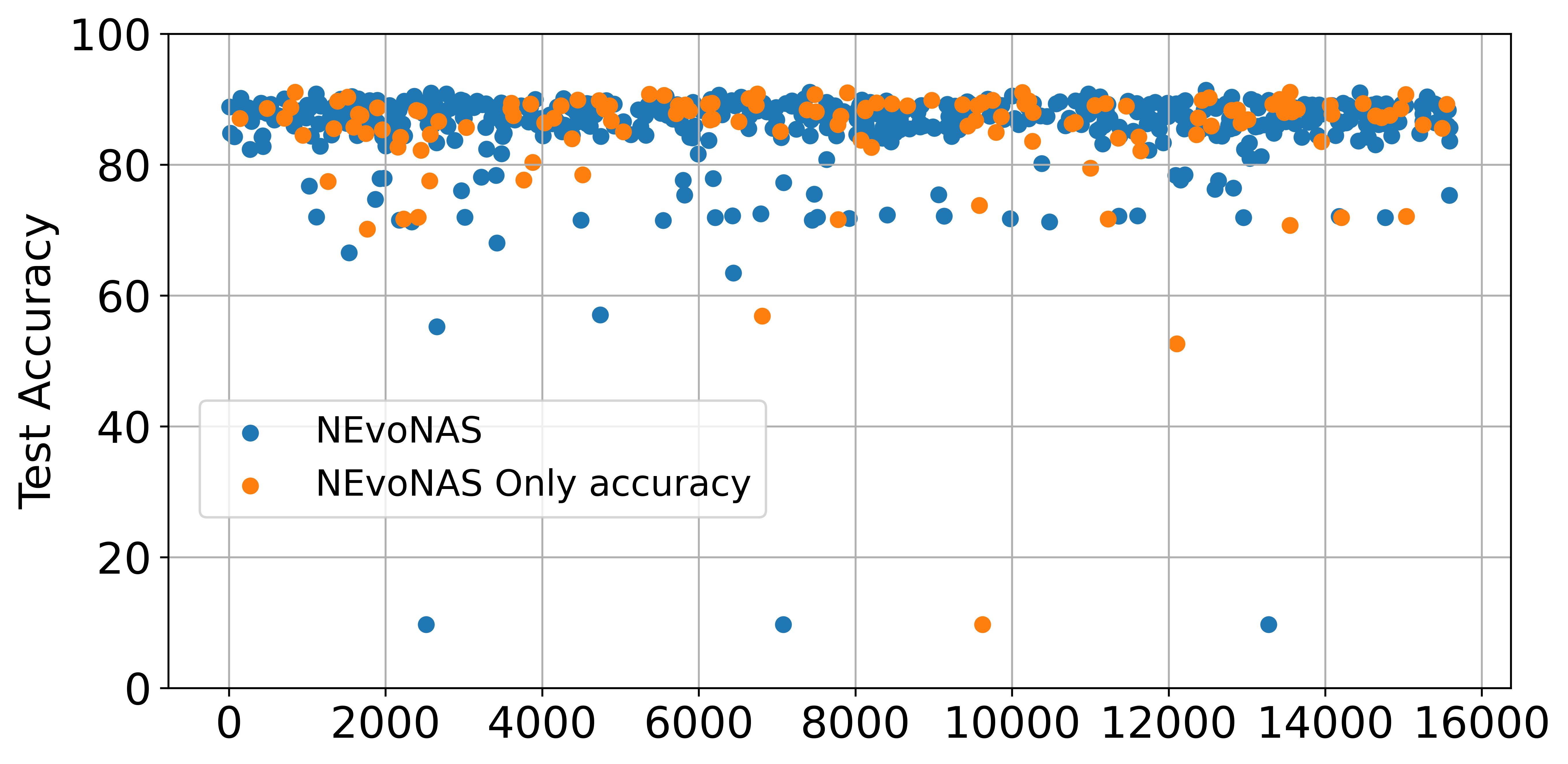}
		\end{center}
		\caption{Visualizing the architecture search space exploration by plotting the
			ground truth accuracies	of all the discovered architectures. The x-axis
			represents all 15,625 architectures in the search space S2
			\cite{Dong2020NAS-Bench-201}, the y-axis represents	the true test
			accuracies of the architectures and the dots represents	all the
			architectures discovered in a single run of NEvoNAS and NEvoNAS using
			only accuracy.
		}
		\label{fig:search_exploration}
	\end{figure}
	}
	
	\section{Futher Analysis}
	We first check the diversity of the population during the architecture search
	by plotting the number of unique individual architectures present in the
	population for each generation of 3 independent runs in
	Figure~\ref{fig:diversity_comparision}. From the
	Figure~\ref{fig:diversity_comparision}(b), we find that when NEvoNAS is applied
	with only accuracy using supernet as an objective, the diversity reduces
	with generation which indicates the convergence to an optimum. As reported in
	\cite{bender2018understanding}, supernet suffers from inaccurate performance
	estimation which results in the convergence to a local optimum. In contrast,
	the diversity does not reduce when NEvoNAS is applied with only novelty as an
	objective as shown in Figure~\ref{fig:diversity_comparision}(c). All these show that
	NEvoNAS maintains a diverse set of individuals in the population during the run
	(as shown in Figure~\ref{fig:diversity_comparision}(a)), owing to the novelty
	search, which helps in avoiding local optimum.
	
	For analyzing the architecture search, we use the search space S2
	\cite{Dong2020NAS-Bench-201} to	visualize the search process as it provides the
	true test accuracies of all the	architectures in the search space. As illustrated
	in Figure~\ref{fig:search_exploration2}, the search process is visualized by
	plotting all the architecures discovered during the single run of the proposed
	algorithm. From the Figure~\ref{fig:search_exploration2}(a), we find that when
	NEvoNAS is applied with only architecture accuracy as objective, the search space
	explored is far less than when NEvoNAS is applied with multi-objective.
	In Figure~\ref{fig:search_exploration2}(b), we plot all the architecures
	discovered when	NEvoNAS is applied with only architecture novelty as objective
	and found that novelty search forces the architecture search to explore the
	architectural search space. But using only novelty as an objective does not
	give good results as it does not have a metric to calculate the quality of the
	architecture. This shows that using novelty search forces NEvoNAS to explore the search
	space which helps in avoiding the local optimum while simultaneously using supernet
	for architecture fitness helps NEvoNAS to explore good quality architectures.
	
	
	\section{Conclusion}
	\label{conclusion}
	The goal of this paper was to mitigate the noisy fitness estimation	nature of the
	supernet which forces NAS methods using supernet to run multiple times to get
	a set of neural architecture solutions. We resolve this by posing the
	NAS problem as a multi-objective problem with two objectives of maximizing the
	architecture novelty (i.e. \textit{novelty search}) and maximizing the architecture
	fitness. This results in a pareto optimal front which provides a set of good quality
	neural architecture	solutions in a single run, thus, reducing the computational
	requirements. We applied NEvoNAS to two different search spaces to show its effectiveness
	in generalizing to any cell-based search space. Experimentally, NEvoNAS reduced the
	search time	of EA-based	search methods significantly while achieving better results
	in S1 search space and state-of-the-art results in S2 search space. We also show that
	using novelty as an objective forces the algorithm to explore the search space by
	maintaining diverse set of individuals in the population in each generation which
	ultimately helps in avoiding the local optimum trap.
	
	An interesting future research direction is to use the novelty search to get a
	list of promising smaller search space areas within the full architectural search
	space.
	
	\section{Acknowledgments}
	This work was supported in part by the Ministry of Science and Technology of Taiwan
	(MOST 110-2628-E-A49-012-, MOST 110-2634-F-A49-006-, and MOST 111-2420-H-369-001-).
	Furthermore, we are grateful to the National Center for High-performance Computing
	for computer time and facilities.
	\comm{
	\subsection{Paper length}
	Papers, excluding the references section, must be no longer than eight pages in length.
	The references section will not be included in the page count, and there is no limit on the length of the references section.
	For example, a paper of eight pages with two pages of references would have a total length of 10 pages.
	{\bf There will be no extra page charges for \confName\ \confYear.}
	
	Overlength papers will simply not be reviewed.
	This includes papers where the margins and formatting are deemed to have been significantly altered from those laid down by this style guide.
	Note that this \LaTeX\ guide already sets figure captions and references in a smaller font.
	The reason such papers will not be reviewed is that there is no provision for supervised revisions of manuscripts.
	The reviewing process cannot determine the suitability of the paper for presentation in eight pages if it is reviewed in eleven.
	}
	{\small
		\bibliographystyle{ieee_fullname}
		\bibliography{references}
	}
	
\end{document}


\newcommand{\comm}[1]{}
	
	
	\title{Supplementary}
	
\maketitle

\section{Discovered Cells in S1}
\begin{figure}[h]
	\qquad
	\subfloat[]{
		\includegraphics[width=\linewidth]{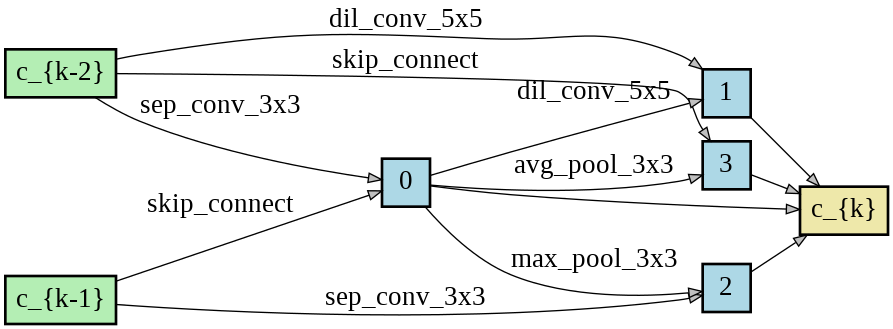}
	}
	\qquad
	\subfloat[]{
		\includegraphics[width=\linewidth]{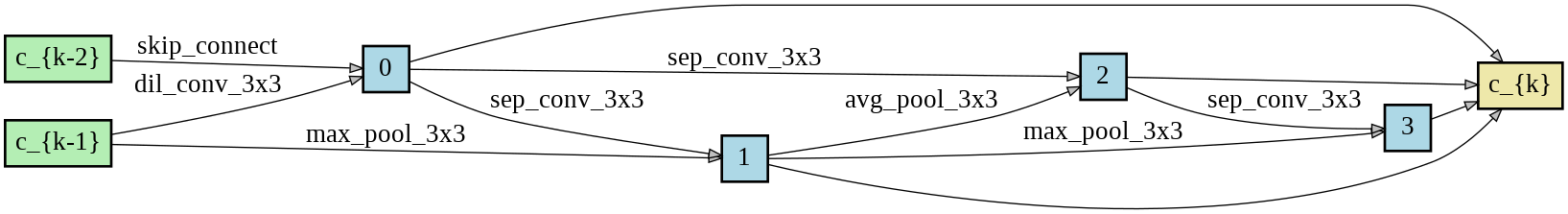}
	}
	\qquad
	\subfloat[]{
		\includegraphics[width=\linewidth]{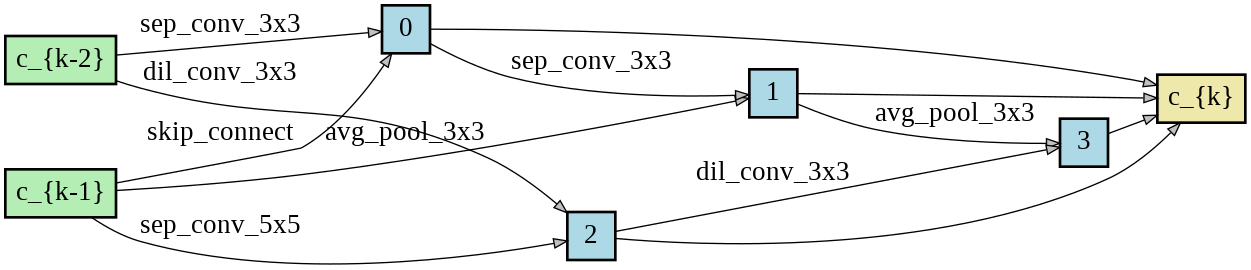}
	}
	\qquad
	\subfloat[]{
		\includegraphics[width=\linewidth]{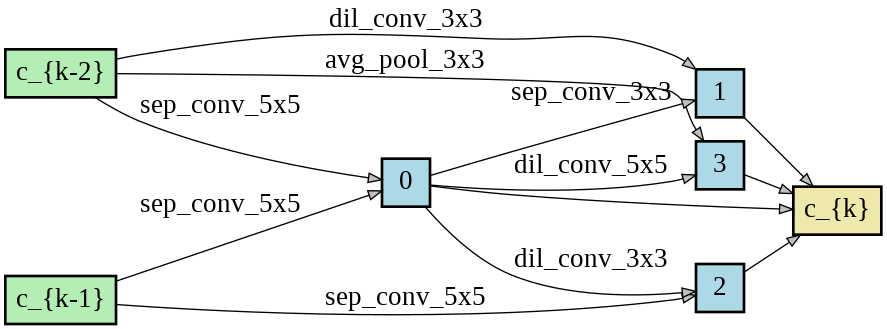}
	}
	\qquad
	
	\caption{Cells discovered 
		by NEvoNAS-C10B (a) Normal cell (b) Reduction cell;
		by NEvoNAS-C10C (c) Normal cell (d) Reduction cell.
	}
	\label{fig:searched_cells_C10}
\end{figure}

\begin{figure}[t!]
	\qquad
	\subfloat[]{
		\includegraphics[width=\linewidth]{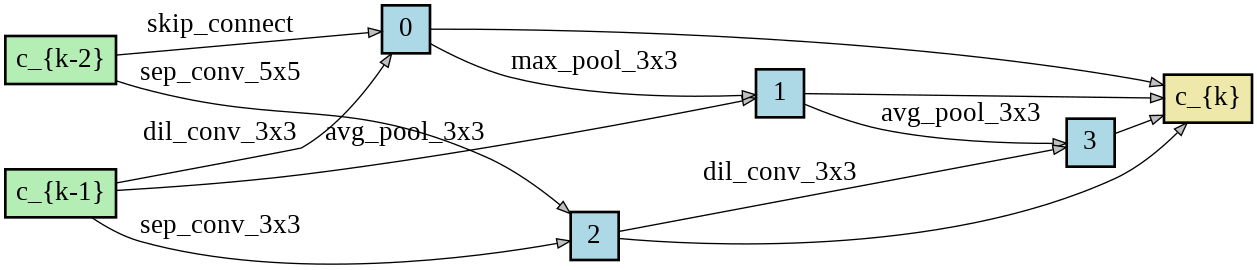}
	}
	\qquad
	\subfloat[]{
		\includegraphics[width=\linewidth]{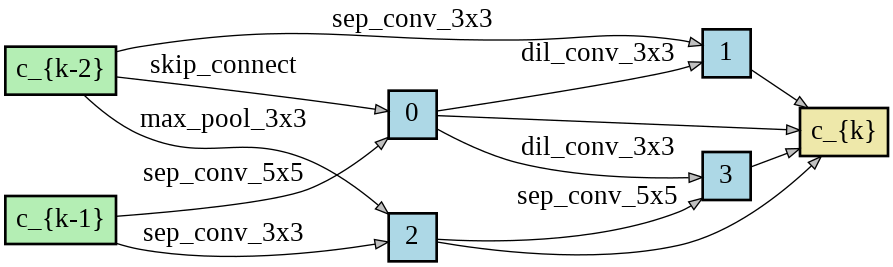}
	}
	\qquad
	\subfloat[]{
		\includegraphics[width=\linewidth]{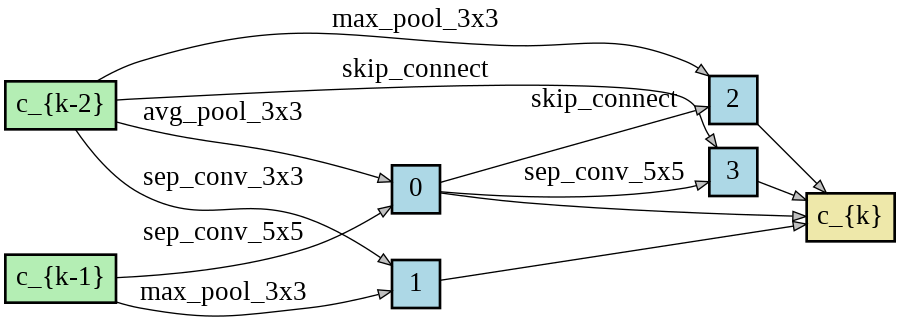}
	}
	\qquad
	\subfloat[]{
		\includegraphics[width=\linewidth]{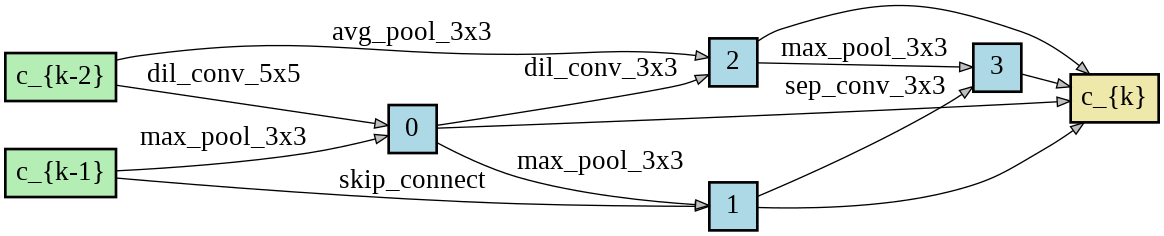}
	}
	
	\caption{Cells discovered 
		by NEvoNAS-C100B (a) Normal cell (b) Reduction cell;
		by NEvoNAS-C100C (c) Normal cell (d) Reduction cell.}
	\label{fig:searched_cells_C100}
\end{figure}

\comm{
\begin{figure*}[b]
	\qquad
	\subfloat[]{
		\includegraphics[width=0.45\linewidth]{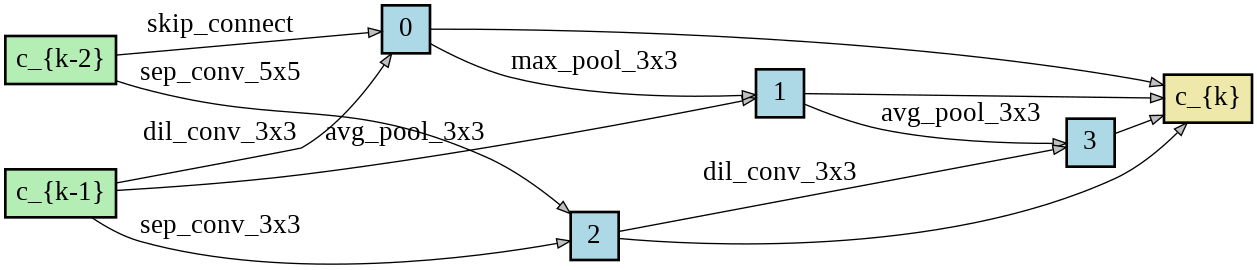}
	}
	\qquad
	\subfloat[]{
		\includegraphics[width=0.45\linewidth]{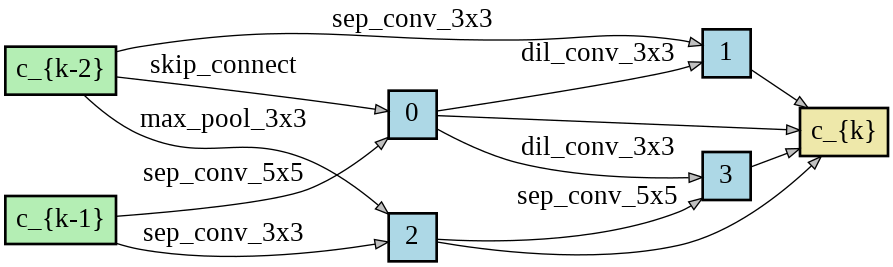}
	}
	\qquad
	\subfloat[]{
		\includegraphics[width=0.45\linewidth]{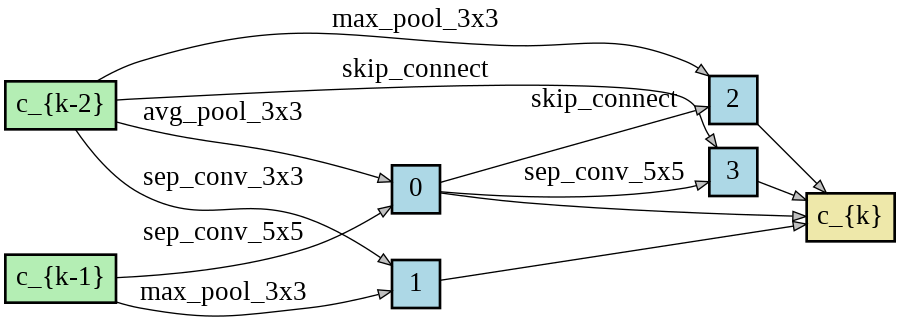}
	}
	\qquad
	\subfloat[]{
		\includegraphics[width=0.45\linewidth]{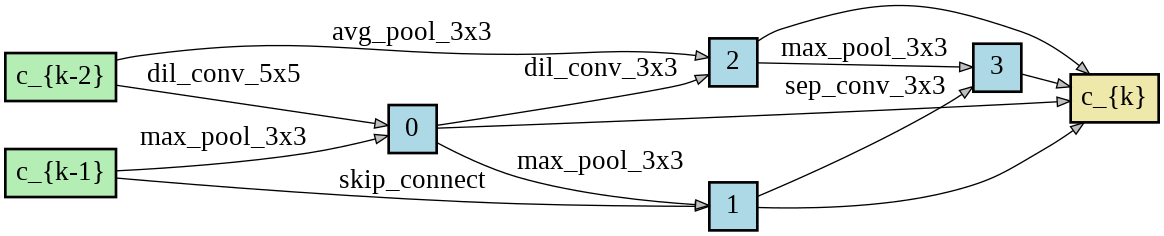}
	}
	
	\caption{Cells discovered 
		by NEvoNAS-C100B (a) Normal cell (b) Reduction cell;
		by NEvoNAS-C100C (c) Normal cell (d) Reduction cell.}
	\label{fig:searched_cells_C100}
\end{figure*}
}
	